%% file: root.tex
\documentclass[letterpaper, 10 pt, conference]{ieeeconf}  

\IEEEoverridecommandlockouts                              

\usepackage{times} 
\usepackage{amsmath} 
\usepackage{amssymb}  
\usepackage{algorithm}
\usepackage{algorithmic}

\usepackage{cite}
\usepackage{amsfonts}

\usepackage{graphicx}
\usepackage{booktabs}
\usepackage{textcomp}
\usepackage{xcolor}
\usepackage{subcaption}
\usepackage{caption}
\let\labelindent\relax
\usepackage{enumitem}

\title{\LARGE \bf
Safety-Constrained Reinforcement Learning with Post-Training Reachability Verification for Robot Navigation
}

\author{Qisong He, Xinmiao Huang, Jinwei Hu, Zhuoyun Li, Yi Dong, Changshun Wu, Xiaowei Huang
\thanks{All authors are with the University of Liverpool, Liverpool, UK.}%
\thanks{Correspondence to: xiaowei.huang@liverpool.ac.uk.}%
}

\begin{document}

\maketitle
\thispagestyle{empty}
\pagestyle{empty}

\begin{abstract}
\input{Secs/0_Abstract}

\end{abstract}

\input{Secs/1_Introduction}

\input{Secs/2_Related_work}

\input{Secs/3_Preliminaries}
\input{Secs/4_Method}
\input{Secs/5_Experiment}
\input{Secs/6_Conclusion}

\bibliographystyle{IEEEtran}
\bibliography{refs}


\end{document}

%% file: Secs/0_Abstract.tex
Safe navigation for mobile robots demands policies that remain reliable under the high-consequence perception uncertainty of cluttered environments. Yet most existing safe reinforcement learning (RL) methods assess safety through average cumulative cost. Such metrics can mask dangerous tail-risk behaviors. To address this, we propose a framework that trains risk-sensitive policies through Conditional Value-at-Risk (CVaR) constrained optimization on an off-policy TD3 backbone and evaluates their safety margins post-training through neural network reachability verification. During training, the policy is optimized under CVaR constraints on cumulative costs, promoting sensitivity to high-cost tail outcomes rather than average behavior alone. After training, we compute action reachable sets under bounded observation uncertainty using Taylor Model analysis, yielding a safety rate metric that quantifies the proportion of evaluated states at which the policy's reachable action set remains within prescribed safety margins. A key finding is that policies trained with CVaR constraints maintain larger safety margins from obstacles across evaluated states. This makes them significantly more amenable to formal reachability verification. Experiments across ten navigation scenarios and six baselines show that our method achieves a 98.3\% success rate, the highest safety verification rate among all compared methods, while revealing that average cost rankings and reachability-based safety rankings can diverge. This indicates that reachability verification captures risks which are missed by empirical cost metrics alone. We further validate our approach on a physical Clearpath Jackal robot, demonstrating successful sim-to-real transfer.

%% file: Secs/1_Introduction.tex
\section{Introduction}
\begin{figure}[t]
    \centering
    \includegraphics[width=0.48\textwidth]{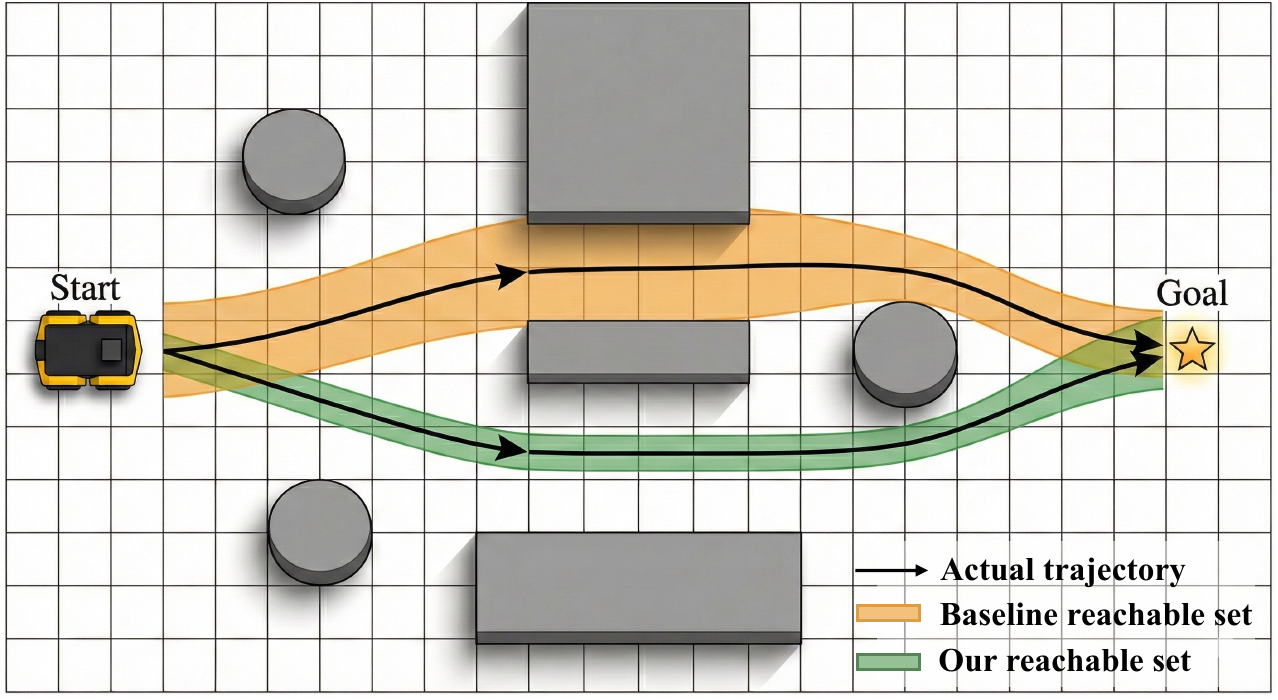}
    \caption{Comparison of action reachable sets under bounded observation
uncertainty. Black lines denote actual trajectories. Orange and green
tubes denote per-step motion envelopes induced by the action reachable
sets of a baseline policy and VIA, respectively.}
    \label{fig:motivation}
\end{figure}

Safe deployment of autonomous mobile robots requires navigation policies that remain reliable not only along a single realized trajectory, but across the full range of behaviors that may arise under perception uncertainty (e.g., sensor noise).
In practice, however, reinforcement learning (RL) policies are typically evaluated by average reward and average cumulative cost. These metrics can mask dangerous tail-risk behaviors. As illustrated in Fig.~\ref{fig:motivation}, two policies may both reach the goal without collision along their nominal trajectories, yet their action reachable sets under bounded observation uncertainty differ substantially. The baseline policy's reachable tube encroaches on obstacles, exposing latent collision risk that conventional metrics fail to capture. This discrepancy motivates combining risk-sensitive training with formal \textit{post-training} verification to assess whether policies deemed safe by training metrics remain verifiable under observation uncertainty.

\begin{figure*}[ht]
    \centering
    \includegraphics[width=1.0\textwidth]{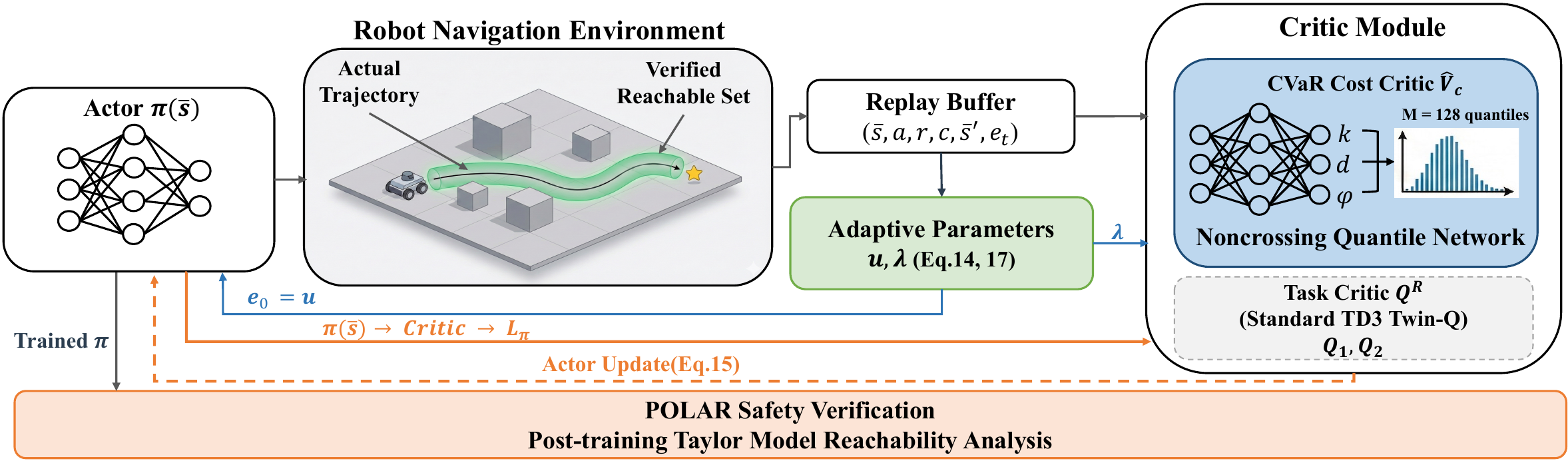}
    \caption{Overview of the VIA framework.}
    \label{fig:framework}
\end{figure*}

A principled approach to safe RL formulates the problem as a constrained Markov decision process (CMDP)~\cite{altman2021constrained}, where the agent maximizes expected return subject to constraints on cumulative costs. Most existing methods enforce these constraints in expectation~\cite{achiam2017constrained,tessler2018reward,sac-lagrangian,ray2019benchmarking}, which limits the overall frequency of unsafe events but provides no control over rare, high-consequence outcomes. Conditional Value-at-Risk (CVaR)~\cite{rockafellar2000optimization} addresses this limitation by constraining the expected cost over the tail of the distribution beyond a chosen quantile, encouraging policies that are robust to high-cost events rather than merely well-behaved on average~\cite{chow2015risk,rahimian2019distributionally,iyengar2005robust}. Nonetheless, existing CVaR-based safe RL methods either rely on distributional assumptions (e.g., Gaussian approximations or trust-region surrogates) that may not hold in practice~\cite{yang2021wcsac,kim2022trc}, or require on-policy updates that limit sample efficiency~\cite{CVaRCPO}. Moreover, a fundamental limitation shared by both expectation-based and CVaR-based approaches is that they provide only statistical training-time signals. A policy with low average cost or favorable CVaR may still produce unsafe actions under the observation perturbations encountered at deployment.
Formal verification methods can address this gap by reasoning over the set of all possible policy outputs. Runtime approaches such as Control Barrier Functions (CBFs)~\cite{ames2016control}, Hamilton-Jacobi (HJ) reachability~\cite{bansal2017hamilton}, and shielding~\cite{alshiekh2018safe} enforce safety online by reasoning over closed-loop system dynamics, which typically requires known analytical models or a differentiable safety function. Polynomial Arithmetic for Reachability (POLAR)~\cite{huang2022polar} targets a different layer of the safety problem. Rather than constraining closed-loop trajectories, it computes guaranteed bounds on a trained network's outputs given interval input perturbations, enabling direct analysis of whether the learned policy's action set remains within prescribed safety margins.

We propose \textbf{VIA} (Verification-Informed safe reinforcement learning for Autonomous navigation), a framework (Fig.~\ref{fig:framework}) that unifies CVaR-constrained policy optimization with post-training reachability-based safety verification for mobile robot navigation. During training, VIA maximizes expected return subject to CVaR-based safety constraints within an off-policy TD3 architecture~\cite{TD3}, providing sample-efficient learning while maintaining sensitivity to high-cost tail events. A key technical challenge in off-policy CVaR optimization is the stable estimation of the Value-at-Risk (VaR) quantile threshold. To address this, we introduce an adaptive VaR update mechanism driven by quantile deviation that reduces oscillation and promotes stable convergence toward the target quantile. After training, we apply POLAR with layered safety boundaries to compute guaranteed bounds on the policy's action outputs given interval input perturbations that represent bounded observation uncertainty. This produces a safety rate metric that quantifies the proportion of evaluated states at which the policy's reachable set remains within prescribed safety margins. Because CVaR-constrained training shapes the policy toward conservative, tightly bounded action distributions, the resulting networks exhibit substantially higher safety rates than expectation-constrained or unconstrained baselines.

Our main contributions are as follows:
\begin{itemize}
    \item We design a CVaR-constrained optimization framework on the TD3 architecture, featuring a distributional cost critic for tail-risk estimation and an adaptive VaR update mechanism that stabilizes quantile convergence.
    \item We develop a safety verification framework combining POLAR reachability analysis with layered safety boundaries and propose safety rate metrics to quantify policy robustness under bounded observation uncertainty, revealing that cost-based and verification-based safety rankings can diverge.
    \item We validate VIA in both simulation and real-world deployment on a Clearpath Jackal robot. The code is publicly available at https://github.com/memory009/VIA
\end{itemize}


%% file: Secs/2_Related_work.tex
\section{Related work}

\subsection{Constrained Safe Reinforcement Learning} 
Within the CMDP framework, Constrained Policy Optimization (CPO)~\cite{achiam2017constrained} performs trust-region policy updates with per-iteration constraint satisfaction, and PCPO~\cite{yang2020projection} and FOCOPS~\cite{zhang2020first} simplify its update procedure while maintaining feasibility guarantees. An alternative line leverages Lagrangian relaxation to convert the constrained problem into a saddle-point formulation: RCPO~\cite{tessler2018reward} introduces a multi-timescale update scheme with convergence guarantees, while SAC-Lagrangian~\cite{sac-lagrangian} and PPO-Lagrangian~\cite{ray2019benchmarking} combine standard off-policy and on-policy backbones with Lagrangian methods. However, all of the above methods constrain only the expected cumulative cost, without explicitly controlling the tail of the cost distribution.

\subsection{CVaR-Based Safe Reinforcement Learning}
To move beyond expectation-based constraints, Chow et al.~\cite{chow2018risk} established a theoretical foundation for CVaR-constrained MDPs with primal-dual policy gradient and actor-critic algorithms. Subsequent approaches differ mainly in how CVaR is estimated and optimized. Worst-Case Soft Actor Critic (WCSAC)~\cite{yang2021wcsac} augments SAC~\cite{haarnoja2018soft} with a safety critic that assumes Gaussian cost returns to derive a closed-form CVaR, an assumption that can fail for heavy-tailed or multi-modal distributions. Trust Region CVaR (TRC)~\cite{kim2022trc} derives a differentiable upper bound on CVaR within a trust-region framework, achieving near-zero violations in an on-policy setting that limits sample efficiency. CVaR-Constrained Policy Optimization (CVaR-CPO)~\cite{CVaRCPO} avoids parametric assumptions via noncrossing quantile networks approximating the full cost distribution, though similarly requiring on-policy updates. We adapt this quantile-based framework to an off-policy TD3 architecture to improve sample efficiency while maintaining stable convergence toward the target safety quantile.

\subsection{Safety Verification and Enforcement for Learned Policies}

Several approaches enforce safety at deployment through runtime intervention, including CBFs~\cite{ames2016control,brunke2022safe}, HJ reachability analysis~\cite{bansal2017hamilton}, and shielding~\cite{alshiekh2018safe}. These methods reason over closed-loop dynamics to constrain or override actions online, but typically require known dynamics models or handcrafted safety certificates. VIA takes a complementary perspective. Rather than enforcing safety through online action correction, we apply neural network reachability verification post-training to compute guaranteed bounds on the policy's outputs under bounded observation uncertainty.


Among existing neural network verification tools~\cite{dutta2019sherlock,ivanov2019verisig,ivanov2021verisig,huang2019reachnn}, we adopt POLAR~\cite{huang2022polar} , which combines Taylor Model (TM) arithmetic with Bernstein polynomial interpolation to produce tight output bounds for networks with general activation functions. However, formal verification remains underexplored in RL for robot navigation: verification cost grows with network size due to TM error accumulation, and existing work focuses on low-dimensional control benchmarks. To our knowledge, VIA is the first to integrate CVaR-constrained policy optimization with neural network reachability verification for robot navigation. We address this gap through lightweight network architectures that enable POLAR verification and a safety rate metric that quantifies policy robustness under bounded observation uncertainty.

%% file: Secs/3_Preliminaries.tex
\section{Preliminaries}

\subsection{Problem Formulation}
\label{sec:problem_formulation}

We consider the safety-constrained reinforcement learning problem modeled as a Constrained Markov Decision Process (CMDP), defined by the tuple $(S, A, P, R, C, \gamma)$, where $S$ is the state space, $A$ is the action space, $P: S \times A \times S \rightarrow [0,1]$ is the transition probability function, $R: S \times A \rightarrow \mathbb{R}$ is the reward function, $C: S \times A \rightarrow \mathbb{R}_{\geq 0}$ is the cost function, and $\gamma \in [0,1)$ is the discount factor.

Let $\pi: S \rightarrow A$ denote a policy. The expected cumulative discounted reward and the episodic cumulative cost are:
\begin{equation}
J_R(\pi) = \mathbb{E}_\pi\left[\sum_{t=0}^{\infty} 
\gamma^t R(s_t, a_t)\right], \quad C^\pi = \sum_{t=0}^{T} C(s_t, a_t)
\end{equation}
where $T$ is the episode length. Note that $C^\pi$ denotes the \emph{undiscounted} episodic cost, which is the random variable constrained in this work. The discounted cost-to-go appears only within the cost critic as a learning surrogate(see the remark in Section~\ref{sec:policy_opt}).

Traditional CMDP methods constrain the expected cumulative cost $\mathbb{E}[C^\pi] \leq b$. However, expectation-based constraints fail to account for the tail risk of the cost distribution. 

For $\alpha \in (0, 1)$, the $\alpha$-level VaR is the $\alpha$-quantile of $C^\pi$. The $\alpha$-level CVaR is defined as:
\begin{equation}
\text{CVaR}_\alpha(C^\pi) = \min_{u \in \mathbb{R}} 
\left\{ u + \frac{1}{1-\alpha}\,\mathbb{E}\left[
\left(C^\pi - u\right)^+\right] \right\}
\label{eq:cvar_def}
\end{equation}
where $(x)^+ = \max(x, 0)$, and the minimizer satisfies 
$u^* = \text{VaR}_\alpha(C^\pi)$. Intuitively, $\text{CVaR}_\alpha$ represents the expected cost in the worst $(1{-}\alpha)$ fraction of episodes.



The objective of this work is to find an optimal policy that maximizes the expected cumulative reward while satisfying the CVaR constraint on cumulative cost:
\begin{equation}
\max_\pi J_R(\pi) \quad \text{s.t.} \quad \text{CVaR}_\alpha(C^\pi) \leq b
\label{eq:cvar_problem}
\end{equation}
where $b > 0$ is the predefined cost threshold. In addition, at deployment the policy acts on a perturbed observation $\tilde{s}$ with bounded error $\|\tilde{s} - s\|_\infty \leq \epsilon$. This uncertainty is not modeled during training but is explicitly handled in the post-training verification phase (Section~\ref{sec:polar_verification}).

The cost function adopts a layered structure commonly used in safety-constrained navigation~\cite{ji2023safety}, with a danger threshold $d_{\text{danger}}$ that triggers a cost signal and a stricter collision threshold $d_{\text{col}} < d_{\text{danger}}$ that terminates the episode:
\begin{equation}
C(s_t, a_t) = \begin{cases} 1 & \text{if } d_{\min}(s_t) 
< d_{\text{danger}} \\ 0 & \text{otherwise} \end{cases}
\end{equation}
where $d_{\min}(s_t)$ denotes the minimum distance to obstacles. 
This layered design provides learning signals before actual collisions occur, encouraging the policy to maintain a safety margin between the two thresholds.

\subsection{TD3 Task Critic and Target Smoothing}
\label{sec:td3_prelim}

We adopt the TD3 twin-critic structure to estimate the expected cumulative reward. Target value is computed as:
\begin{equation}
y_R = r + \gamma \min_{i=1,2} 
Q_{i,\psi'}^{R}(s', \tilde{a}')
\end{equation}
where $r = R(s_t, a_t)$ is the reward, and $\tilde{a}' = \pi^{\text{tar}}(s') + \xi$ is the smoothed target action with clipped noise $\xi \sim \text{clip}(\mathcal{N}(0, \sigma), -c, c)$. The task critics $Q_{i,\psi}^{R}$ are updated by minimizing the mean squared error $\mathcal{L}_Q = \sum_{i=1}^{2} \mathbb{E}[(Q_i^R(s, a) - y_R)^2]$.

\subsection{POLAR-based Safety Verification}
\label{sec:polar_verification}

We integrate POLAR~\cite{huang2022polar}, in the optimized form of Dong et al~\cite{dong2024reachability}, to perform post-training safety assessment of the learned policy under bounded observation uncertainty via Taylor Model (TM) reachability analysis.

Assuming a feedforward policy network with known activation functions, POLAR computes over-approximated reachable sets by propagating TMs through the network. A TM is a pair $(p, I)$, where $p(\mathbf{z})$ is a polynomial over symbolic variables and $I$ is an interval remainder bounding the approximation error. Given a state $s$ and uncertainty bound $\epsilon$, the perturbed input is:

\begin{equation}
\tilde{s}_i = s_i + \epsilon \cdot z_i, \quad z_i \in [-1, 1]
\end{equation}
The TM propagates through each layer. Linear transformations apply as $(p', I') = W \cdot (p, I) + \text{bias}$, while activation functions are approximated through Bernstein polynomial approximation:
\begin{equation}
\sigma(x) \in p_\sigma(x) + I_\sigma, \quad \forall x \in (p', I')
\end{equation}
After propagating through all layers, the output TM defines the action reachable set:
\begin{equation}
\mathcal{A}(s, \epsilon) = \left\{ a : a \in p_a(\mathbf{z}) 
+ I_a,\, \mathbf{z} \in [-1,1]^n \right\}
\label{eq:action_reachable_set}
\end{equation}
This set provides \emph{guaranteed bounds} on all possible actions the policy may produce given observation uncertainty bounded by $\epsilon$.




%% file: Secs/4_Method.tex
\section{Method}

VIA operates in two phases (Fig.~\ref{fig:framework}). In the training phase, we adapt CVaR-constrained optimization to an off-policy TD3 framework through state augmentation for CVaR estimation, a distributional cost critic with noncrossing quantile regression, adaptive VaR parameter updates, and a stabilized Lagrangian formulation (Algorithm~\ref{alg:via}). In the verification phase, we apply POLAR reachability analysis to the trained policy to provide safety assessment under bounded observation uncertainty.

\subsection{Phase I: CVaR-Constrained Off-Policy Training}

\subsubsection{State Augmentation for CVaR Estimation}
\label{sec:state_aug}
To enable CVaR-constrained optimization, we extend the state space to track cumulative cost information~\cite{CVaRCPO}. The augmented state is defined as $\bar{s}_t = (s_t, e_t)$, where $e_t$ serves as a dynamic cost budget variable initialized at each episode start as $e_0 = u$, with $u$ being an adaptive parameter approximating the $\alpha$-quantile of the cumulative cost distribution. This augmentation is necessary because CVaR is a property of the cumulative cost distribution: without $e_t$, a critic observing only $s_t$ cannot distinguish states reached under different cost histories. The budget evolves within each episode following:
\begin{equation}
e_{t+1} = \frac{e_t - C(s_t, a_t)}{\gamma}
\label{eq:budget_update}
\end{equation}
This recursive update captures accumulated cost history in a Markovian manner, and $e_t$ additionally serves as the threshold for selecting quantiles in the CVaR computation.

\input{Tabs/Pretend_code}

\input{Tabs/Baseline_result_all}

\subsubsection{Distributional Cost Critic}
\label{sec:cost_critic}
To accurately estimate CVaR, we need to learn the distribution of cumulative cost rather than just its expectation. We employ a distributional cost critic $Q^C_\phi$ that outputs $M$ quantiles $\{q_i\}_{i=1}^M$ with uniformly spaced fractions $\tau_k = k/M$. 



A key challenge in quantile regression is the crossing problem, where predicted quantiles violate monotonicity $q_i \leq q_j$ for $i < j$. We address this using a noncrossing quantile logit network\cite{zhou2020non} with the following parameterization:
\begin{equation}
q_i(\bar{s}, a) = k(\bar{s}, a) \cdot \varphi_i(\bar{s}, a) + d(\bar{s}, a)
\end{equation}
where $k = \text{softplus}(f_k(h)) > 0$ ensures a positive slope, $d = f_d(h)$ is an unconstrained bias term, and $\varphi_i = \sum_{j=1}^{i} \text{softmax}(f_\varphi(h))_j$  is computed using cumulative softmax summation to guarantee monotonicity. Here $h$ denotes the hidden representation from the shared network layers.


Using this quantile representation, we estimate the CVaR-related cost value $\hat{V}_C(\bar{s}_t, a_t)$, defined as the conditional mean of predicted quantiles exceeding the budget variable $e_t$:
\begin{equation}
\hat{V}_C(\bar{s}_t, a_t) = \frac{\sum_{i=1}^{M} q_i \cdot \mathbb{I}(q_i \geq e_t)}{\sum_{i=1}^{M} \mathbb{I}(q_i \geq e_t)}
\end{equation}

The cost critic is trained using the Huber quantile regression loss:
\begin{equation}
\mathcal{L}_C = \frac{1}{M^2} \sum_{i=1}^{M} \sum_{j=1}^{M} \rho_{\hat{\tau}_i}^\kappa (\delta_{ij})
\label{eq:cost_critic_loss}
\end{equation}
where $\delta_{ij} = c + \gamma q_j(\bar{s}', a') - q_i(\bar{s}, a)$ is the temporal difference error, $\hat{\tau}_i = (\tau_{i-1}+\tau_i)/2$ are the quantile midpoints, and $\rho_{\hat{\tau}_i}^\kappa$ is the Huber quantile loss~\cite{dabney2018distributional}, which smooths 
gradients near zero error compared to the standard pinball loss.
\subsubsection{VaR Parameter Adaptive Update}
\label{sec:var_update}

A key challenge in off-policy CVaR optimization is tracking the VaR quantile without relying on on-policy cost critic evaluations, which become unreliable when the critic is trained on replay buffer data from earlier policies. We address this with a gradient-free update rule based on empirical exceedance probability. The VaR is initialized as $u^0$, the empirical $\alpha$-quantile of costs collected during warm-up. At each epoch $k$, the empirical exceedance probability is computed as:
\begin{equation}
\hat{P}(C \geq u^k) = \frac{1}{N}\sum_{n=1}^{N}
\mathbb{I}\left(C^{(n)} \geq u^k\right)
\end{equation}
where $C^{(n)}$ is the undiscounted cumulative cost of the $n$-th episode and $N$ is the number of episodes in epoch $k$. The VaR parameter is then updated as:
\begin{equation}
u^{k+1} = u^k + \beta_u \cdot \left[\hat{P}(C \geq u^k) 
- (1-\alpha)\right]
\label{eq:var_update_rule}
\end{equation}
where $\beta_u > 0$ is the learning rate and the bracketed term drives $u$ toward the true $\alpha$-quantile. This update is gradient-free with respect to the cost critic parameters, making it directly compatible with off-policy learning.

\subsubsection{CVaR-Constrained Policy Optimization}
\label{sec:policy_opt}

We optimize the policy through Lagrangian relaxation of the CVaR-constrained objective:
\begin{equation}
\min_\lambda \max_\pi \, 
\mathbb{E}[Q^R_\psi(\bar{s}, \pi(\bar{s}))] 
- \lambda \cdot \left(\mathbb{E}[\hat{V}_C(\bar{s}, 
\pi(\bar{s}))] - b + u\right)
\end{equation}
where $\lambda \geq 0$ is the Lagrange multiplier, $b$ is the cost threshold.

\textit{Task Critic.} We adopt the standard TD3 twin-critic architecture described in Sec.~\ref{sec:td3_prelim}.

\textit{Actor.} The actor loss is defined as:
\begin{equation}
\mathcal{L}_\pi = -\mathbb{E}\left[Q^R_\psi(\bar{s}, \pi(\bar{s}))\right] + \lambda \cdot \mathbb{E}\left[\hat{V}_C(\bar{s}, \pi(\bar{s}))\right]
\label{eq:actor_loss}
\end{equation}
Following the TD3 delayed update schedule, the actor and all target networks are synchronized every 2 critic updates via exponential moving average with coefficient $\rho \in (0,1)$.

\textit{Lagrange Multiplier.} The Lagrange multiplier is updated based on the empirical constraint slack:
\begin{equation}
\lambda^{k+1} = \text{Proj}_{[0, \lambda_{\max}]}\left[\lambda^k - \beta_\lambda \cdot (b - u^k - \bar{C}_k)\right]
\label{eq:lagrangian_update_rule}
\end{equation}
where $\bar{C}_k = \frac{1}{N}\sum_{n=1}^{N} C^{(n)}$ is the average undiscounted episodic cost over epoch $k$, $\beta_\lambda > 0$ is the learning rate and $\lambda_{\max}$ is the projection bound. While the on-policy CVaR-CPO uses the distributional cost value $\hat{V}_C(\bar{s}_0)$ in this update, our off-policy cost critic may be queried on states from earlier policies in the replay buffer, making it less reliable as a constraint signal. We therefore use the empirical mean $\bar{C}_k$ from current-epoch rollouts together with $u^k$, which already encodes tail information as the $\alpha$-quantile tracker.

\textit{Remark (discounted vs.\ undiscounted cost).} The cost critic and budget recursion (Eqs.~\eqref{eq:budget_update}, \eqref{eq:cost_critic_loss}) operate on the \emph{discounted} cost-to-go for stable off-policy TD learning, whereas the VaR and Lagrange updates (Eqs.~\eqref{eq:var_update_rule}, \eqref{eq:lagrangian_update_rule}) use \emph{undiscounted} episodic costs. The constraint enforced is thus the CVaR of the undiscounted $C^\pi$ in Eq.~\eqref{eq:cvar_problem}, with the discounted critic acting as a per-state surrogate that mildly downweights later violations.


\subsection{Phase II: Post-Training Safety Verification}
\label{sec:polar_verification_mehod}
While the CVaR constraint encourages tail-risk sensitivity during training, it provides only statistical guarantees over the cost distribution. We complement this with post-training safety assessment using POLAR reachability analysis. Verification is performed on the augmented state $\bar{s} = (s, e_t)$ with $e_t$ fixed to the converged VaR parameter $u$, consistent with the initial budget assignment at episode start. This assessment leverages the action reachable set $\mathcal{A}(s, \epsilon)$ defined in Section~\ref{sec:polar_verification} to determine whether the policy's outputs remain within prescribed safety margins.

\subsubsection{Safety Verification Criterion} 
Given the action reachable set $\mathcal{A}(s, \epsilon) = [\underline{a}, \overline{a}]$ computed by POLAR at state $s$, we assess whether all actions within the reachable set lead to collision-free motion. We evaluate actions sampled from the reachable set bounds, including the interval endpoints and midpoint per action dimension, yielding 9 combinations spanning the extremal regions of the 2D action space (linear and angular velocity). For each sampled action, we simulate the robot's motion over the control interval $\Delta t$ under a differential-drive kinematic model, discretizing the trajectory into multiple time steps and checking the robot's swept area at each step against the obstacle map. A state $s$ is \emph{assessed as safe} if no collision is detected across all sampled actions and all time steps along their resulting trajectories:
\begin{equation}
\text{Safe}(s) = \bigwedge_{j=1}^{9} \left( \min_{p \in 
\mathcal{S}(s, a_j, \Delta t)} d(p, \mathcal{O}) > 
d_{\text{col}} \right), \quad a_j \in \mathcal{A}(s, \epsilon)
\label{eq:safety_criterion}
\end{equation}
where $\mathcal{S}(s, a, \Delta t)$ denotes the swept area of the robot executing action $a$ from state $s$ over time interval $\Delta t$ under the differential-drive kinematic model, $d(p, \mathcal{O})$ is the minimum distance from point $p$ to obstacle region $\mathcal{O}$, and $d_{\text{col}}$ is the collision threshold defined in Section~\ref{sec:problem_formulation}. We note that the formal guarantee extends to the action bounds certified by POLAR. The subsequent collision check is a \emph{per-timestep} sampled robustness test over extremal actions of $\mathcal{A}(s,\epsilon)$, not a formal proof of closed-loop safety.

\subsubsection{Safety Rate Metric} Based on per-state safety assessment, we define the Safety Rate as the proportion of trajectory states satisfying the safety criterion:
\begin{equation}
\text{Safety Rate} = \frac{1}{T} \sum_{t=1}^{T} 
\mathbb{I}\left[\text{Safe}(s_t)\right]
\label{eq:safety_rate}
\end{equation}
where $T$ is the total number of evaluated states across all trajectories. Unlike empirical collision rates that only capture actual failures, the Safety Rate quantifies the policy's robustness margin under observation uncertainty, providing a verification-informed metric that complements the CVaR-based training objective.

\input{Figs/Trajectories}

%% file: Tabs/Pretend_code.tex
\begin{algorithm}[t]
\caption{VIA: CVaR-Constrained Off-Policy Training}
\label{alg:via}
\textbf{Input:} parameterized policy $\pi_\theta$, task critic network for rewards $Q^R_\psi$,
cost critic network for cost $Q^C_\phi$, confidence level $\alpha$, 
cost threshold $b$, discount factor $\gamma$, learning rates $\beta_\pi, \beta_u, \beta_\lambda$.

\begin{algorithmic}[1]
\STATE \textbf{Warm-up Phase (VaR Initialization):}

Collect $N$ episodes using initial policy (no training);\\
$u^0 \leftarrow \alpha$-quantile of $\{C^{\text{ep}}_i\}_{i=1}^N$

\STATE \textbf{Initialization:}
Policy $\theta=\theta_0$; value $\psi=\psi_0$, $\phi=\phi_0$;
Replay buffer $\mathcal{B} \leftarrow \emptyset$; VaR $u \leftarrow u^0$; Lagrangian $\lambda \leftarrow \lambda_0$; initial condition $\bar{s}_0=(s_0,e_0)=(s_0,u^0)$.

\FOR{epoch $k = 1, \ldots, K$}
    \STATE $\mathcal{C}_k \leftarrow \{\}$ \COMMENT{Episode costs for this epoch}
    
    \FOR{episode $n = 1, \ldots, N$}
        \STATE Initialize $e_0 \leftarrow u^k$
        \FOR{$t = 0, 1, \ldots, T$}
            \STATE $a_t \sim \pi_\theta(\cdot|\bar{s}_t)$ where $\bar{s}_t = (s_t, e_t)$
            \STATE $s_{t+1} \sim P(\cdot|s_t, a_t)$, $e_{t+1} \leftarrow (e_t - C(s_t, a_t)) / \gamma$
            \STATE $\mathcal{B} \leftarrow \mathcal{B} \cup \{(\bar{s}_t, a_t, R_t, C_t, \bar{s}_{t+1})\}$
        \ENDFOR
        \STATE $\mathcal{C}_k \leftarrow \mathcal{C}_k \cup \{C^{\text{ep}}_n\}$


    \ENDFOR
    \STATE Perform $U$ off-policy updates by sampling from $\mathcal{B}$: \\
    \hspace{1em} update $Q^C_\phi$ via Eq.~\eqref{eq:cost_critic_loss}, $\pi_\theta$ via Eq.~\eqref{eq:actor_loss}
    

    \STATE Update $u^{k+1}$ via VaR update rule Eq.~\eqref{eq:var_update_rule}
    \STATE Update $\lambda_{k+1}$ via Lagrangian update rule Eq.~\eqref{eq:lagrangian_update_rule}
\ENDFOR
\end{algorithmic}
\end{algorithm}

%% file: Tabs/Baseline_result_all.tex
\begin{table*}[ht]
\centering
\normalsize
\caption{Experimental Results on Robot Navigation Task}
\label{tab:main_results}
\begin{tabular}{lccccccc}
\toprule
\textbf{Method} & \textbf{Success} $\uparrow$ & \textbf{Collision} $\downarrow$ & \textbf{Avg Cost} $\downarrow$ & \textbf{Safety (Succ.)} $\uparrow$ & \textbf{Safety (Coll.)} $\uparrow$ & \textbf{Overall Safety} $\uparrow$ \\
\midrule
TD3 & 76.7$\pm$4.7 & 15.0$\pm$7.6 & -- & 98.9$\pm$1.3 & 85.2$\pm$5.3 & 98.4$\pm$0.8 \\
TD7 & 78.0$\pm$7.5 & 12.0$\pm$7.5 & -- & 97.3$\pm$0.3 & 84.7$\pm$3.2 & 95.1$\pm$1.0 \\
\midrule
SAC-Lagrangian & 64.0$\pm$4.9 & 36.0$\pm$4.9 & 3.52$\pm$6.42 & 97.4$\pm$2.4 & 86.4$\pm$4.8 & 94.3$\pm$2.5 \\
RCPO & 90.0$\pm$8.9 & 4.0$\pm$4.9 & \textbf{1.17$\pm$1.46} & 98.5$\pm$0.3 & 91.7$\pm$0.2 & 97.4$\pm$0.3 \\
\midrule
WCSAC & 88.0$\pm$7.5 & 12.0$\pm$7.5 & 1.79$\pm$4.29 & 97.5$\pm$0.7 & 88.1$\pm$4.4 & 95.6$\pm$1.7 \\
CVaR-CPO & 76.0$\pm$4.9 & 24.0$\pm$4.9 & 3.22$\pm$3.16 & 96.5$\pm$1.9 & 87.2$\pm$2.7 & 94.1$\pm$1.5 \\
\textbf{VIA (Ours)} & \textbf{98.3$\pm$4.9} & \textbf{1.7$\pm$4.9} & 1.52$\pm$1.53 & \textbf{99.6$\pm$0.6} & \textbf{94.4$\pm$0.2} & \textbf{99.1$\pm$0.6} \\
\bottomrule
\end{tabular}
\end{table*}

%% file: Figs/Trajectories.tex
\begin{figure}[t]
    \centering

    \begin{subfigure}[t]{0.32\linewidth}
        \centering
        \includegraphics[width=\linewidth]{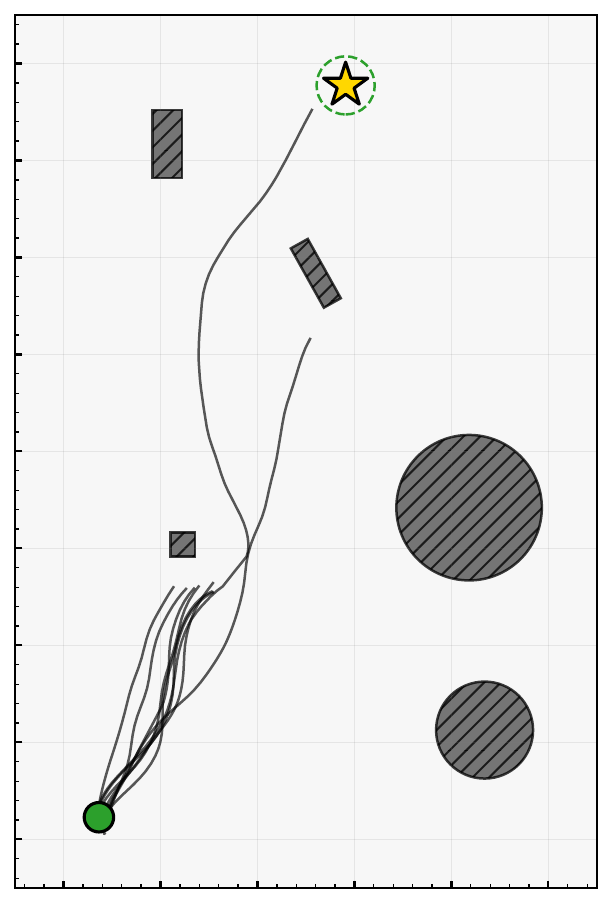}
        \caption{TD3}
    \end{subfigure}\hfill
    \begin{subfigure}[t]{0.32\linewidth}
        \centering
        \includegraphics[width=\linewidth]{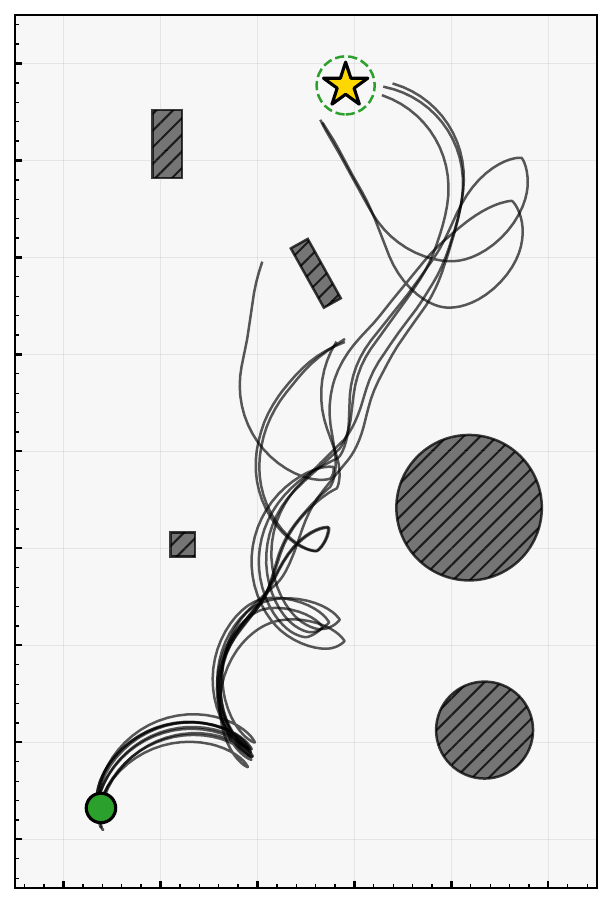}
        \caption{TD7}
    \end{subfigure}\hfill
    \begin{subfigure}[t]{0.32\linewidth}
        \centering
        \includegraphics[width=\linewidth]{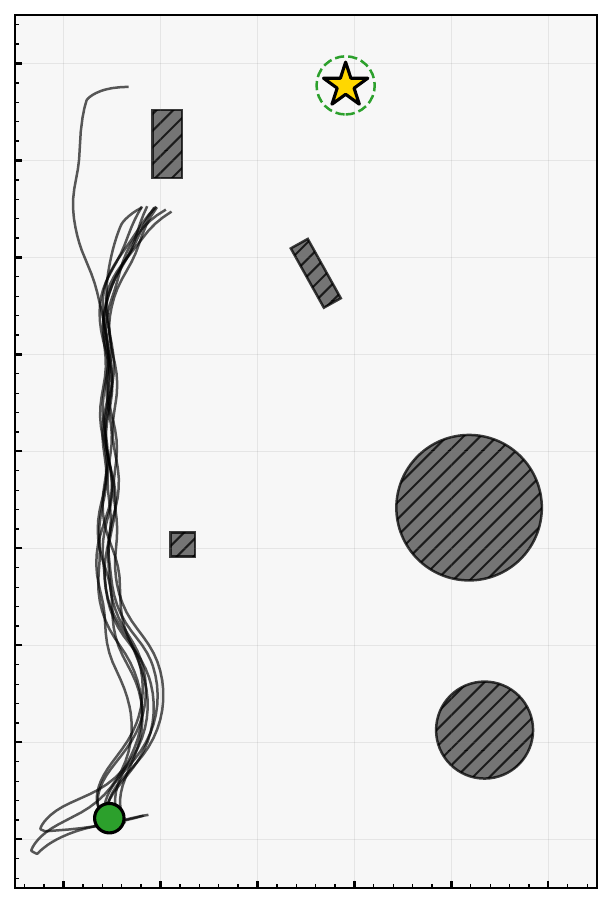}
        \caption{SAC-Lagrangian}
    \end{subfigure}

    \vspace{2mm}

    \begin{subfigure}[t]{0.32\linewidth}
        \centering
        \includegraphics[width=\linewidth]{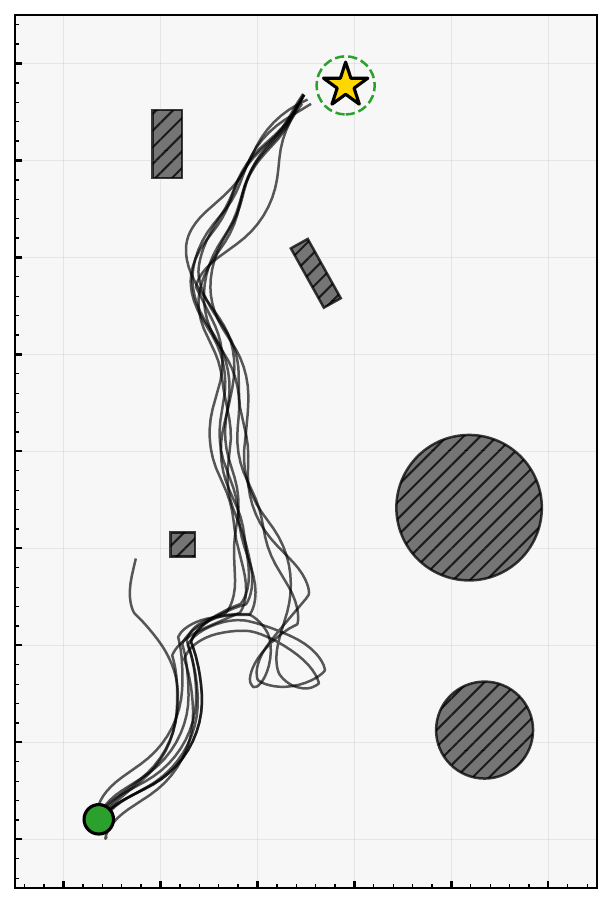}
        \caption{RCPO}
    \end{subfigure}\hfill
    \begin{subfigure}[t]{0.32\linewidth}
        \centering
        \includegraphics[width=\linewidth]{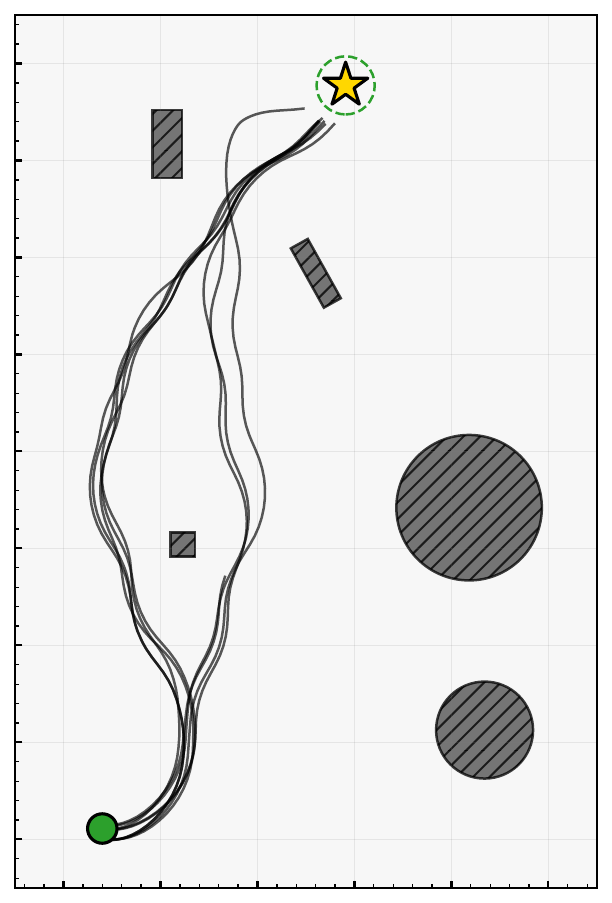}
        \caption{WCSAC}
    \end{subfigure}\hfill
    \begin{subfigure}[t]{0.32\linewidth}
        \centering
        \includegraphics[width=\linewidth]{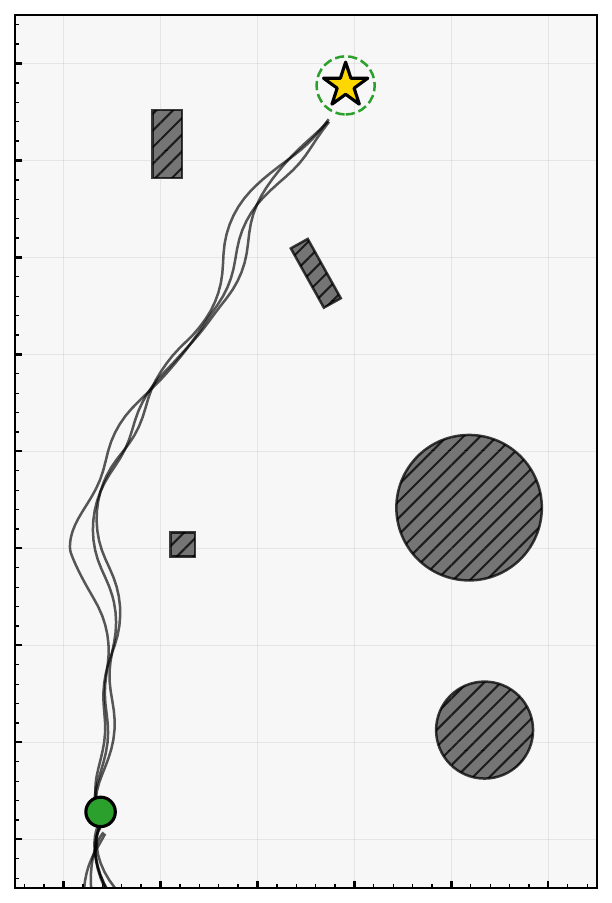}
        \caption{CVaR-CPO}
    \end{subfigure}

    \vspace{2mm}

    \begin{subfigure}[t]{0.32\linewidth}
        \centering
        \includegraphics[width=\linewidth]{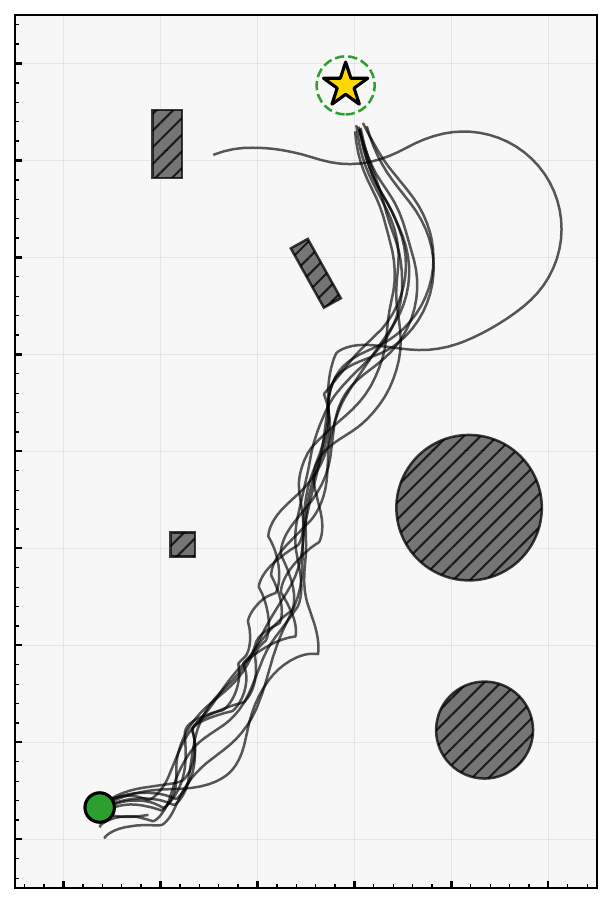}
        \caption{VIA-0.1}
    \end{subfigure}\hfill
    \begin{subfigure}[t]{0.32\linewidth}
        \centering
        \includegraphics[width=\linewidth]{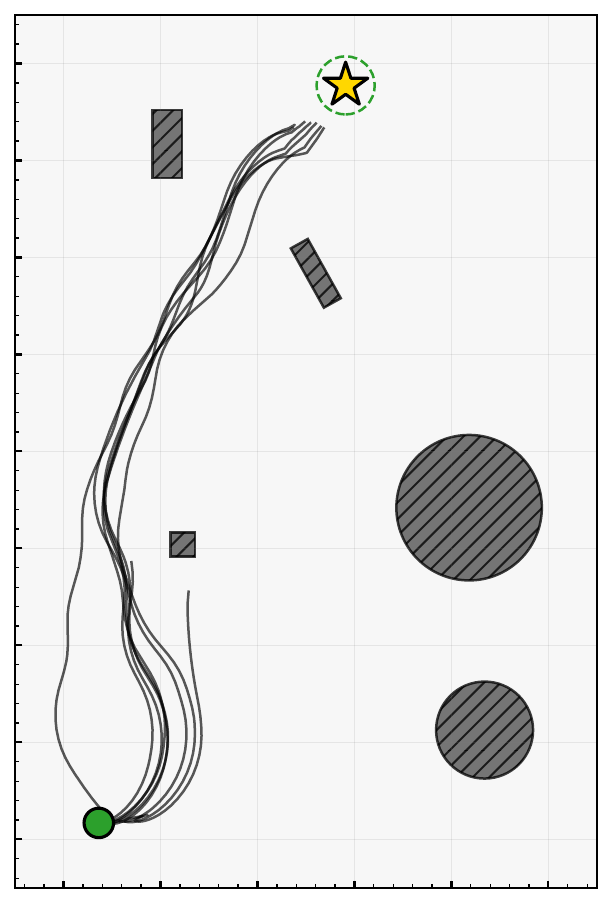}
        \caption{VIA-0.5}
    \end{subfigure}\hfill
    \begin{subfigure}[t]{0.32\linewidth}
        \centering
        \includegraphics[width=\linewidth]{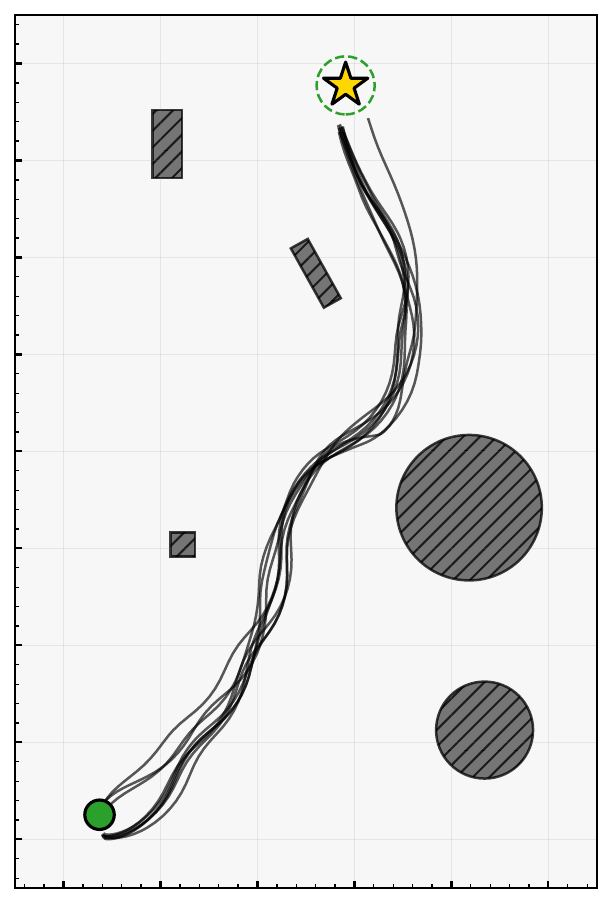}
        \caption{VIA-0.9}
    \end{subfigure}

    \caption{Navigation trajectories for different constraint formulations. Unconstrained RL: (a), (b); Expectation-constrained RL: (c), (d); CVaR-constrained RL: (e), (f); VIA with CVaR $\alpha \in \{0.9, 0.5, 0.1\}$: (g), (h), (i).}
    \label{fig:trajectories}
\end{figure}

%% file: Secs/5_Experiment.tex
\section{Experiments}
\label{sec:experiments}
In this section, we conduct comprehensive experiments to evaluate the proposed VIA algorithm in a robot navigation task and address the following questions:
\begin{enumerate}[label=\textbf{Q\arabic*}:, leftmargin=2em]
    \item Does CVaR-based risk formulation improve navigation and safety performance compared to unconstrained and expectation-based methods?
    \item Can reachability verification reveal safety risks that conventional cost metrics fail to capture?
    \item How does the CVaR confidence level $\alpha$ affect policy conservatism and task performance?
    \item Can the learned policy maintain its safety performance when deployed on a physical robot?
\end{enumerate}

\subsection{Experimental Setup}

\subsubsection{Environment} We evaluate all methods in a Gazebo-based simulation~\cite{DRL-robot-navigation} with a $10\text{m} \times 10\text{m}$ arena containing 8 cylindrical obstacles (4 fixed + 4 random). A TurtleBot3 robot equipped with 180° LiDAR navigates to goal positions while avoiding collisions. The state space is 25-dimensional (20 LiDAR bins, goal distance, goal angle, previous actions), and the action space is 2-dimensional (linear and angular velocity). For VIA and CVaR-CPO, the policy input is the 26-dimensional augmented state $\bar{s}_t = (s_t, e_t)$ including the cost budget (Section~\ref{sec:state_aug}). Each episode terminates upon collision, goal arrival, or exceeding 300 steps (timeout).

\subsubsection{Safety Thresholds} We adopt the layered cost structure from Section~\ref{sec:problem_formulation}, with collision threshold $d_{\text{col}} = 0.4$\,m and danger threshold $d_{\text{danger}} = 0.5$\,m. All constrained methods share the same binary per-step cost: $c_t = 1$ if the agent enters the danger threshold or collision, and $c_t = 0$ otherwise.

\subsubsection{Training} All methods are trained for 100 epochs with 70 episodes per epoch using consistent hyperparameters. Discount factor $\gamma=0.99$, learning rate $10^{-4}$, and hidden dimension 26 (to enable POLAR verification). All methods share the same reward function, cost signal, and termination conditions. CVaR-specific parameters include risk level $\alpha=0.9$ and cost threshold $b=10$. Method-specific hyperparameters follow the recommended values in the respective original papers. Training is performed on a laptop with an NVIDIA RTX 3060 GPU.


\subsubsection{POLAR Verification} We apply POLAR reachability analysis with observation error $\epsilon=0.01$, consistent with the target policy smoothing noise scale used in TD3 training. For each state, we compute the action reachable set and verify safety by sampling actions from the reachable set bounds and checking collision-free motion over the control interval. Verification is CPU-bound and runs with 10 parallel threads on an AMD EPYC 7452 processor, taking on average 2.5\,s per state ($\approx$3.8\,min per trajectory).


\subsubsection{Evaluation Metrics} All methods are trained with 5 random seeds and evaluated on 10 fixed start--goal pairs in each epoch. We report mean $\pm$ std across seeds. Task performance is measured by \emph{Success Rate}, \emph{Collision Rate}, and \emph{Average Cost}. For verification-based metrics, we evaluate the per-state safety criterion (Eq.~\eqref{eq:safety_criterion}) at every state along recorded trajectories. \textit{Safety (Succ.)} and \textit{Safety (Coll.)} report the fraction of states assessed as safe (Eq.~\eqref{eq:safety_rate}) over successful and collision episodes, respectively, and \textit{Overall Safety} covers all evaluated states.

\input{Tabs/Compare_4}

\subsection{Baselines}
To isolate the effect of constraint formulation, all methods share the same TD3 backbone, network architecture, and hyperparameters.

\subsubsection{Unconstrained RL} TD3~\cite{TD3} optimizes task reward without safety constraints. TD7~\cite{TD7} extends TD3 with state-action embeddings for improved sample efficiency.

\subsubsection{Expectation-Constrained RL} SAC-Lagrangian~\cite{sac-lagrangian} augments the actor loss with a Lagrangian penalty on expected cumulative cost, adapted from SAC to TD3. RCPO~\cite{tessler2018reward} penalizes the reward function based on constraint violation, adapted from on-policy to TD3.

\subsubsection{CVaR-Constrained RL} WCSAC~\cite{yang2021wcsac} estimates CVaR via Gaussian approximation, with SAC replaced by TD3. CVaR-CPO~\cite{CVaRCPO} uses quantile regression with adaptive VaR updates, similarly adapted to TD3.

\subsection{Main Results (Q1, Q2)}
We evaluate all methods on the robot navigation task and report reachability-based safety metrics under bounded observation uncertainty. Results are summarized in Table~\ref{tab:main_results}.

\textbf{Q1: Safety constraint formulation significantly affects performance.}
The effectiveness of safety constraints depends critically on how risk is formulated. Expectation-based methods constrain average behavior but leave tail risk unaddressed, which manifests as the highest collision rate among constrained methods in SAC-Lagrangian (36.0\%). Trajectories are shown in Fig.~\ref{fig:trajectories}(c),~\ref{fig:trajectories}(d). CVaR-based formulations explicitly penalize worst-case outcomes, and VIA achieves the best success rate (98.3\%) and lowest collision rate (1.7\%) among all methods (Fig.~\ref{fig:trajectories}(e),~\ref{fig:trajectories}(f),~\ref{fig:trajectories}(i)).

\textbf{Q2: Average cost alone is an insufficient safety indicator.}
The most revealing comparison is between RCPO and VIA. RCPO achieves the lowest average cost among constrained methods (1.17), which under conventional evaluation would suggest it is the safest policy. However, reachability verification reveals otherwise. VIA achieves higher Safety (Succ.) (99.6\% vs.\ 98.5\%) and Safety (Coll.) (94.4\% vs.\ 91.7\%) despite a higher average cost (1.52). Trajectories are shown in Fig.~\ref{fig:trajectories}(d) and~\ref{fig:trajectories}(i). This indicates that average cost captures nominal performance, while the safety rate captures robustness to bounded observation uncertainty.


\subsection{Ablation Study (Q3)}We investigate the effect of the CVaR confidence level $\alpha$ on VIA's performance. As defined in Section~\ref{sec:problem_formulation}, $\text{CVaR}_\alpha$ measures the expected cost in the worst $(1{-}\alpha)$ fraction of episodes. A higher $\alpha$ focuses on a smaller, more extreme tail, yielding a more conservative policy. Note that this convention follows ~\cite{CVaRCPO}. Some works~\cite{yang2021wcsac} use $1{-}\alpha$ in place of $\alpha$, under which our $\alpha{=}0.9$ corresponds to their $\alpha{=}0.1$.

As shown in Table~\ref{tab:baseline_comparison}, as $\alpha$ increases from 0.1 to 0.9, both success rate and safety metrics improve consistently. In robot navigation, worst-case avoidance and task completion are aligned objectives. A policy that reliably avoids dangerous situations is more likely to complete long-horizon trajectories. Fig.~\ref{fig:trajectories}(g)--(i) reflects this directly,
with $\alpha=0.9$ producing the most consistent paths and narrowest reachable action sets under verification.

\input{Tabs/Robot}

\subsection{Real-World Validation (Q4)}
\begin{figure}[h]
    \centering
    \includegraphics[width=0.49\columnwidth]{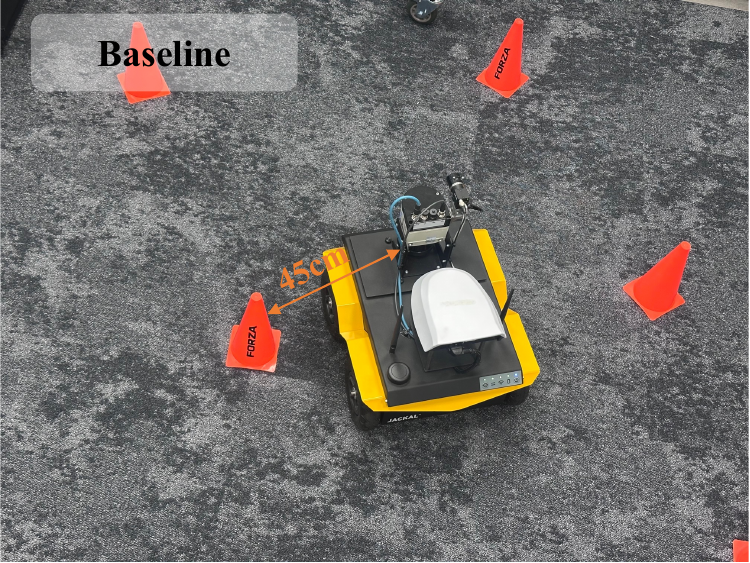}
    \includegraphics[width=0.49\columnwidth]{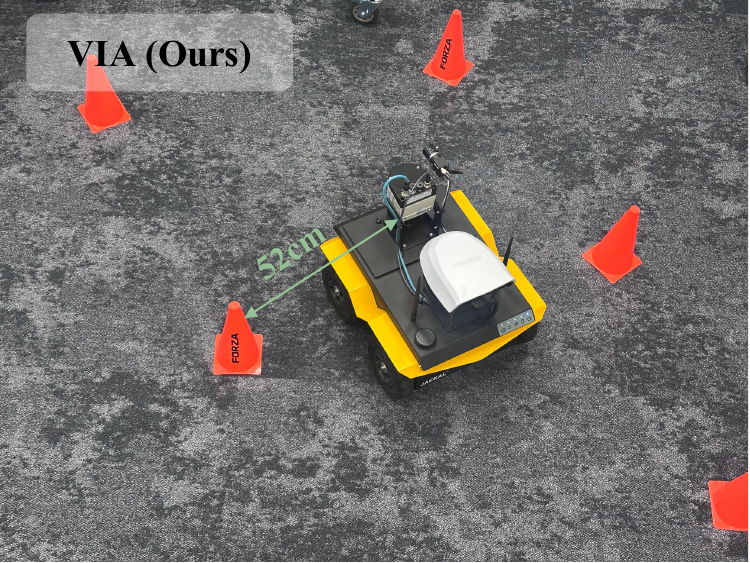}
    \caption{Real-world deployment on a Clearpath Jackal robot. 
(a) Baseline policy violates the danger threshold ($d < 0.5$m). 
(b) VIA ($\alpha=0.9$) maintains clearance outside the danger threshold throughout navigation.}
    \label{fig:real_robot}
\end{figure}

To validate sim-to-real transfer, we deploy the trained VIA policy ($\alpha=0.9$) on a physical Clearpath Jackal robot in a laboratory environment matching the simulation layout. Table~\ref{tab:sim2real} reports performance across both settings over 500 real world trials. The real-world success rate decreases from 98.3\% to 90.0\%, reflecting the increased difficulty of physical deployment including sensor noise and actuation delays. Despite this task-level gap, the overall safety rate remains nearly identical between simulation and real-world deployment (99.1\% vs.\ 99.2\%), indicating that VIA's conservative, CVaR-constrained policy maintains verified safety margins even under real-world perturbations. Reachable set widths also remain consistent across domains ($v$: 0.019 vs.\ 0.020\,m/s; $\omega$: 0.087 vs.\ 0.099\,rad/s), confirming that the policy's action distribution does not degrade during transfer. Fig.~\ref{fig:real_robot} illustrates the robot maintaining clearance from obstacles during navigation, consistent with the verification results established in simulation.

%% file: Tabs/Compare_4.tex
\begin{table}[t]
    \caption{Comparison in different CVaR $\alpha$}
    \centering
    \resizebox{0.9\columnwidth}{!}{%
    \label{tab:baseline_comparison}
    \begin{tabular}{lccc}
        \toprule
        Metric 
        & $\alpha=0.1$ 
        & $\alpha=0.5$ 
        & $\alpha=0.9$  \\
        \midrule
        Success Rate (\%)
        & 86.0$\pm$4.9
        & 90.0$\pm$0.0
        & \textbf{98.3$\pm$4.9} \\
        Collision Rate (\%)
        & 12.0$\pm$4.0
        & 10.0$\pm$0.0
        & \textbf{1.7$\pm$4.9} \\
        Safety (Succ.) (\%)
        & 99.5$\pm$0.8
        & 99.6$\pm$0.1
        & \textbf{99.6$\pm$0.6} \\
        Safety (Coll.) (\%)
        & 87.3$\pm$5.0
        & 92.1$\pm$0.6
        & \textbf{94.4$\pm$0.2} \\
        Overall Safety (\%) 
        & 98.7$\pm$0.5
        & 97.6$\pm$0.4
        & \textbf{99.1$\pm$0.6} \\
        \bottomrule
    \end{tabular}%
    }
\end{table}

%% file: Tabs/Robot.tex
\begin{table}[t]
\caption{Sim-to-Real Transfer Results (VIA, $\alpha=0.9$)}
\label{tab:sim2real}
\centering
\resizebox{\columnwidth}{!}{%
\begin{tabular}{lccc}
\toprule
Metric & Simulation & Real-world \\
\midrule
Success Rate (\%) & 98.3 $\pm$ 4.9 & 90.0 $\pm$ 6.3 \\
Collision Rate (\%) & 1.7 $\pm$ 4.9 & 8.0 $\pm$ 4.0 \\
Safety(Succ.) (\%)& 99.6 $\pm$ 0.6 & 99.9 $\pm$ 0.1 \\
Safety(Coll.) (\%)& 94.4 $\pm$ 0.2 & 91.0 $\pm$ 2.2 \\
Overall Safety (\%) & 99.1 $\pm$ 0.6 & 99.2 $\pm$ 0.6\\
Reachable Set Width $v$ (m/s) & 0.019 $\pm$ 0.003 & 0.020 $\pm$ 0.003 \\
Reachable Set Width $\omega$ (rad/s) & 0.087 $\pm$ 0.018 & 0.099 $\pm$ 0.014 \\
\bottomrule
\end{tabular}%
}
\end{table}

%% file: Secs/6_Conclusion.tex
\section{Conclusion}


In this work, we proposed VIA, a safe reinforcement learning framework that integrates CVaR-constrained optimization into an off-policy TD3 architecture for robot navigation and complements training-time risk control with post-training POLAR reachability analysis. Experiments show that reachability-based safety evaluation can reveal risk discrepancies that conventional cost metrics do not capture. Real-world deployment on a Clearpath Jackal robot demonstrates successful sim-to-real transfer. A current limitation is that Taylor Model arithmetic requires compact network architectures, restricting the policy complexity that can be verified. Future work will explore GPU-accelerated reachability methods and network compression to scale verification to larger policies, as well as integrating verification signals into the training loop, e.g., using reachable-set width or per-state verification outcomes as an additional cost signal, to enable closed-loop safety optimization.

\section*{Acknowledgments}
This work is partially funded by the European Union (under grant agreement ID 101212818). Views and opinions expressed are however those of the author(s) only and do not necessarily reflect those of the European Union or European Health and Digital Executive Agency (HADEA). Neither the European Union nor the granting authority can be held responsible for them. 
This work is partially supported by Innovate UK through AI-PASSPORT under Grant 10126404. This work was awarded a grant by the AI Security Institute (AISI) via the Alignment Project
(Rare-Event Estimation in Large Language Models via Subset Simulation) and funded by EPSRC.
Yi's contribution is partially supported through the Royal Society international exchanges programme and in part by the Engineering and Physical Sciences Research Council, through funding from RAi UK [EP/Y009800/1]. 